\documentclass[journal]{IEEEtran}

\usepackage{xcolor,soul,framed} 

\colorlet{shadecolor}{yellow}
\usepackage[pdftex]{graphicx}
\graphicspath{{../pdf/}{../jpeg/}}
\DeclareGraphicsExtensions{.pdf,.jpeg,.png}

\usepackage[cmex10]{amsmath}
\usepackage{array}
\usepackage{mdwmath}
\usepackage{mdwtab}
\usepackage{graphicx}
\usepackage{eqparbox}
\usepackage{url}
\usepackage{amssymb}
\usepackage{booktabs}
\usepackage{multirow}

\begin{document}
\bstctlcite{IEEEexample:BSTcontrol}
    \title{UVCPNet: A UAV-Vehicle Collaborative Perception Network for 3D Object Detection}
\author{Yuchao~Wang,
        Peirui~Cheng,
        Pengju~Tian,
        Ziyang~Yuan,
        Liangjin~Zhao,
        Jing~Tian, \IEEEmembership{Member,~IEEE},
        Wensheng~Wang, \IEEEmembership{Member,~IEEE},
        Zhirui~Wang, \IEEEmembership{Member,~IEEE},
        Xian~Sun, \IEEEmembership{Senior Member,~IEEE}
\thanks{This work was supported by the National Nature Science Foundation of China under Grant 62331027 and Grant 62076241, and supported by the Strategic Priority Research Program of the Chinese Academy of Sciences, Grant No. XDA0360300.
\textit{(Corresponding author: Zhirui~Wang.)}}
\thanks{Yuchao Wang and Xian Sun are with the Aerospace Information Research Institute, Chinese Academy of Sciences, Beijing 100190, China, also with the Key Laboratory of Network Information System Technology (NIST), Aerospace Information Research Institute, Chinese Academy of Sciences, Beijing 100190, China, also with the University of Chinese Academy of Sciences, Beijing 100190, China, and also with the School of Electronic, Electrical and Communication Engineering, University of Chinese Academy of Sciences, Beijing 100190, China (e-mail: wangyuchao22@mails.ucas.ac.cn; sunxian@aircas.ac.cn). }
\thanks{Peirui Cheng, Zhirui Wang, Liangjin Zhao, Wensheng Wang and Jing Tian are with the Aerospace Information Research Institute, Chinese Academy of Sciences, Beijing 100094, China, and also with the Key Laboratory of Network Information System Technology (NIST), Aerospace Information Research Institute, Chinese Academy of Sciences, Beijing 100190, China (e-mail: chengpr@aircas.ac.cn; zhirui1990@126.com; zhaolj004896@aircas.ac.cn; wangws@aircas.ac.cn; tianjing@aircas.ac.cn).}.
}


\maketitle

\begin{abstract}


With the advancement of collaborative perception, the role of aerial-ground collaborative perception, a crucial component, is becoming increasingly important. The demand for collaborative perception across different perspectives to construct more comprehensive perceptual information is growing. However, challenges arise due to the disparities in the field of view between cross-domain agents and their varying sensitivity to information in images. Additionally, when we transform image features into Bird’s Eye View (BEV) features for collaboration, we need accurate depth information. To address these issues, we propose a framework specifically designed for aerial-ground collaboration. First, to mitigate the lack of datasets for aerial-ground collaboration, we develop a virtual dataset named V2U-COO for our research. Second, we design a Cross-Domain Cross-Adaptation (CDCA) module to align the target information obtained from different domains, thereby achieving more accurate perception results. Finally, we introduce a Collaborative Depth Optimization (CDO) module to obtain more precise depth estimation results, leading to more accurate perception outcomes. We conduct extensive experiments on both our virtual dataset and a public dataset to validate the effectiveness of our framework. Our experiments on the V2U-COO dataset and the DAIR-V2X dataset demonstrate that our method improves detection accuracy by 6.1\% and 2.7\%, respectively.
\end{abstract}


%
\IEEEpeerreviewmaketitle


\section{Introduction}

\IEEEPARstart {C}ollaborative perception  \cite{liu2020who2com,han2023collaborative,hu2022where2comm} achieves efficient and comprehensive research methods for environmental monitoring and data collection by integrating the advantages of multiple sensor networks. This technology enables the coverage of a broad geographic area while providing detailed ground-level information to address the issues of occlusion and small object detection from high altitudes, thereby significantly enhancing the accuracy and efficiency of monitoring efforts. The technology supports a variety of application scenarios, including agricultural precision management, urban planning, and disaster response. This is of great significance for intelligent and automated decision support systems \cite{chen2022hierarchical}.

In the task of collaborative perception between intelligent agents, BEV feature maps can unify different field of view information into feature maps at the same level, which has been widely used and has important significance in collaborative perception. By converting the feature information of optical images into the BEV feature space, information from various perspectives can be uniformly presented on the same feature level. However, the majority of research \cite{xu2023bridging,cui2022coopernaut,vadivelu2021learning,wang2020v2vnet} on collaborative perception utilizing BEV feature maps operates under an overly simplistic assumption: that all intelligent agents are homogenous or have a small height gap. Specifically, this entails that all agents are equipped with identical sensor models and share the same detection model. However, with the advancement of computer vision and collaborative perception technology, the role of aerial-ground collaboration has become increasingly significant. As drones are integrated into collaborative networks for joint tasks, the participating intelligent agents become heterogeneous. Exploring aerial-ground collaboration based on the foundation of homogenous agent collaborative perception research presents us with two primary challenges.

\begin{figure}
  \begin{center}
  \includegraphics[width=3.5 in]{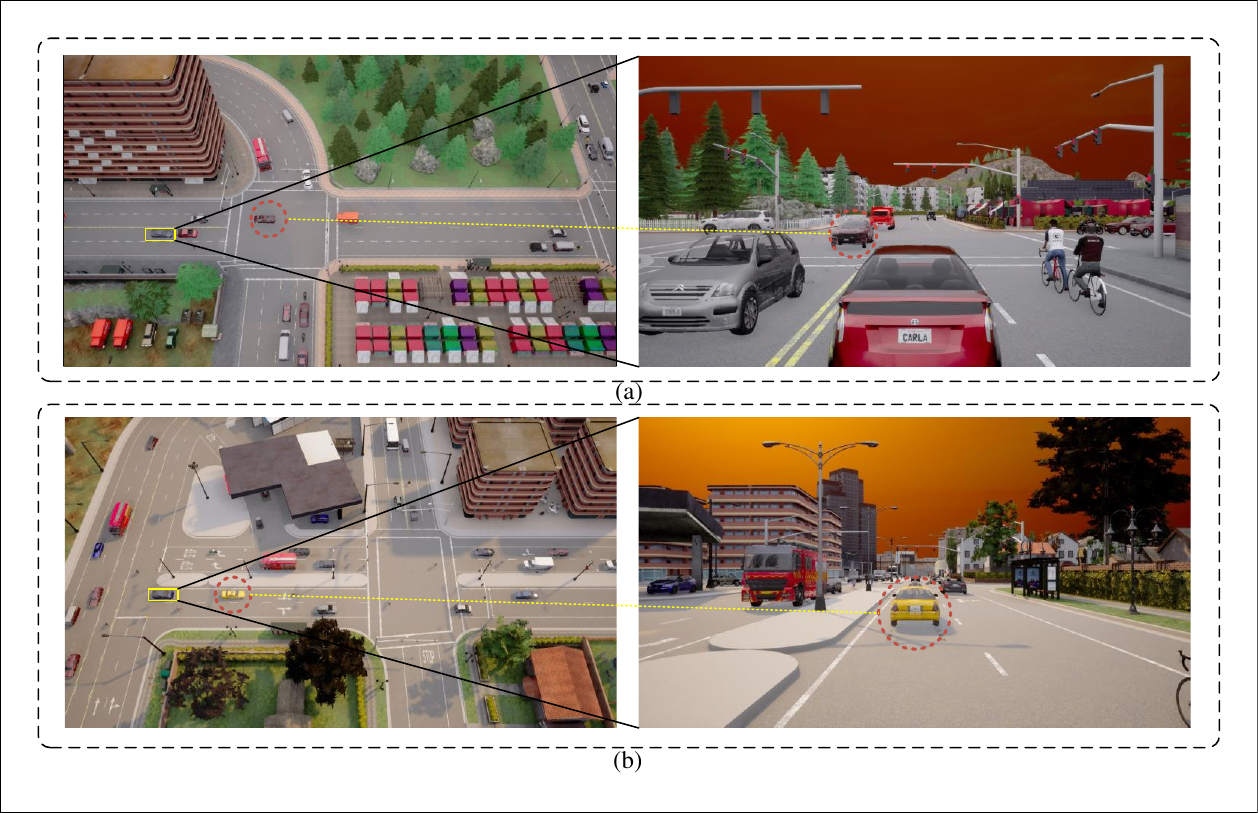}\\
  \caption{ Left images and represent the scene images observed from the UAV's perspective, while the right images and respectively depict the vehicle images annotated in left images. It can be observed that there is a significant disparity in the observed field of view between the aerial and ground domains.}\label{circuit_diagram}
  \end{center}
\end{figure}

The first challenge pertains to the issue of cross-domain collaborative perception among heterogeneous agents. In the study of heterogeneous agents, there are considerable disparities in the visual information among agents, as well as varying degrees of sensitivity to information in images, as depicted in Fig. 1. This results in the generation of BEV feature maps for each domain being more inclined to include information from areas sensitive to that particular domain, while neglecting important information from other domains. This inevitably leads to the creation of a domain gap. Such a gap hinders the effective fusion of perceptual information by the agents and reduces their capabilities for collaborative detection. The second challenge pertains to the inaccuracy of depth estimation encountered when generating  BEV feature maps from optical images across different domains. The generation of BEV feature maps relies on the accurate estimation of depth and scene information within optical images. However, the depth range of targets relative to agents varies across different domains, indicating that the estimated depth values differ when applied within two distinct networks. This leads to inconsistencies in depth estimation for optical images under different agents. When utilizing these inconsistent depth information for BEV perspective transformation, the scene presented in the BEV feature space will exhibit errors in the relative positions of targets observed by agents across different domains.

To address the challenges mentioned above, we propose a collaborative task network architecture named UVCPNet, a framework specifically designed for aerial-ground collaboration. To tackle the first challenge, we introduce a Cross-Domain Cross-Adaptation (CDCA) module aimed at aligning the BEV feature maps generated from different domains. After generating BEV feature maps on cross-domain agents, this module facilitates the cross-adaptation of BEV features' relevance among agents, thereby enhancing the features obtained by each agent. By aligning the features acquired by each agent in the BEV feature map space, we ultimately achieve the effective integration of information from both aerial and ground perspectives. To response the second challenge, given that accurate depth estimation is essential for collaborative perception of BEV feature maps at each agent, we introduce a Collaborative Depth Optimization (CDO) module. This module optimizes depth estimation using Conditional Random Fields (CRF) \cite{Lan_Yang_Yu_Wu_Alvarez_Anandkumar_2023,Lan_Yu_Choy_Radhakrishnan_Liu_Zhu_Davis_Anandkumar_2021,Liu_Chunhua_Shen, yuan2203new}. By transforming the optical image information of each agent to a unified scale through a camera parameter network, it integrates the depth estimation information obtained from cross-domain agents with the contextual pixel information of optical images. Utilizing the pixel information of optical images to optimize the depth estimation results, this approach enhances the accuracy of depth estimation. To assess the effectiveness of our network architecture and further advance research in aerial-ground collaboration, we introduce a large-scale aerial-ground collaborative dataset named V2U-COO. The dataset contains a large number of optical images of UAV and vehicle cooperation, which will be described in detail later.

In summary, our contributions can be summarized as follows:

1. We introduce the pioneering UVCPNet framework, a multi-view aerial-ground collaborative framework based on BEV. Notably, this framework represents the first task framework custom-designed specifically for aerial-ground collaboration.

2. To address the challenge of cross-domain information fusion during the collaborative process, we designed a module named CDCA. This module cleverly overcomes the barriers to effective collaboration by promoting seamless integration of information across different domains.

3. To enhance the precision of depth estimation in collaborative perception scenarios, we introduce the CDO module. Utilizing CRF, this module autonomously monitors and refines depth estimation information, ensuring an improvement in accuracy throughout the collaborative process.

4. In a groundbreaking initiative, we meticulously curated the V2U-COO dataset, custom-tailored for aerial-ground collaboration. This dataset represents a pioneering contribution and stands as the most significant dataset specifically designed for collaborative perception between vehicles and drones.

\section{RELATED WORK}

In this section, we provide an overview of related work, focusing primarily on advancements in aerial-ground collaboration, cross-domain perception, and depth estimation.

\subsection{Aerial-Ground Coordination}

Research on aerial-ground collaboration is still in its preliminary stages, with most studies concentrating on exploring multi-stage perspectives. For instance, AGCG \cite{sun2024agcg} proposed a visual servoing approach based on a binocular localization model, adjusting the alignment between the unmanned aerial vehicle (UAV) and the target in the image to determine the target's position. Utilizing multi-feature fusion recognition and projection imaging techniques, this method enhances the positioning accuracy of UAVs and unmanned ground vehicles (UGVs) in outdoor environments concerning moving targets. However, these methods have not yet proposed a specific aerial-ground collaboration research framework and have only undergone simulated testing, lacking support from actual datasets to validate the accuracy of the methods.

In recent years, with the development of drone technology, many high-quality datasets (such as VisDrone \cite{zhu2018visdrone,du2019visdrone,cao2021visdrone}, HIT-UAV \cite{suo2023hit}, DroneVehicle \cite{sun2022drone}, etc.) have emerged, containing only images from the UAV perspective. The recently released MAVREC \cite{dutta2023multiview} dataset provides images from both ground and aerial perspectives, but lacks three-dimensional annotations and coordinate transformations between UAVs and UGVs, thus only usable as an object detection dataset, incapable of achieving target-level aerial-ground collaborative perception.
In summary, current aerial-ground collaboration research mainly focuses on object detection from a single perspective, followed by its extension to other perspectives to supplement information. However, we aspire to integrate information from both "aerial" and "ground" domains to obtain more comprehensive collaborative perception results. This will aid in more accurately understanding and addressing targets in complex environments.

\subsection{Cross-Domain Perception}

As collaborative perception tasks mature in the field of autonomous driving, significant progress has been made in collaborative work among homogeneous agents. For example, HM-ViT \cite{HM-ViT} demonstrates efficient collaborative capabilities by fusing point cloud and optical image data. Additionally, in the context of drone swarm collaboration, When2com \cite{when2com} proposed a grouping-based cooperative method that successfully reduces the bandwidth requirements. These studies mainly focus on collaborative perception tasks within the domain and do not address challenges in cross-domain scenarios.

To address challenges in cross-domain collaborative perception, some high-quality datasets such as DAIR-V2X \cite{yu2022dairv2x} and V2X-Sim \cite{li2022v2x} have emerged, along with proposed collaborative methods. However, most cross-domain collaborative methods \cite{LIDAR1,LIDAR2,LIDAR3,LIDAR4,LIDAR5} still rely on Lidar data. For instance, FFNet \cite{yu2023ffnet} solves the problem of asynchronous timing between vehicles and infrastructure using point cloud data, while TransIFF \cite{TransIFF} utilizes instance-level features to enhance the robustness of feature fusion. However, due to the large storage space occupied by Lidar data, there is a desire to directly use optical images for cross-domain collaborative perception to reduce storage space usage.

In summary, existing research has made significant progress in collaborative perception tasks within the field and partially addressed challenges in cross domain scenarios. Afterwards, we need to further explore and develop methods that directly utilize optical images for cross domain collaborative perception.

\subsection{Depth Estimation}

Depth estimation tasks \cite{depth1,DEPTH2,DEPTH3}, as essential auxiliary tasks in 3D visual detection \cite{3D1,3D2,3D3,3D4}, play a crucial role in the modeling process, where accurate depth values help avoid distortions. With the continuous maturity of depth estimation algorithms, some new methods have made remarkable progress. For example, News-CRF treats depth estimation tasks as pixel-level classification problems, making them applicable to CRF applications and achieving advanced depth estimation results.

However, most depth estimation in the collaborative perception domain still relies on Lidar data. For depth estimation from optical images, it primarily relies on the depth estimation scheme proposed in LSS \cite{philion2020lift}. In collaborative perception, CoCa3D \cite{CoCa3D} fully utilizes the depth collaboration probability of each intelligent agent to modulate depth values. Although these methods have improved the accuracy of depth estimation to some extent, there is still room for optimization, especially in fully utilizing collaborative information.
We hope to fully utilize collaborative information by employing CRF to optimize depth estimation schemes for more accurate feature information. This may include interacting information among multiple intelligent agents and jointly optimizing depth estimation tasks with other relevant tasks to obtain more comprehensive and accurate depth estimation results.

\section{METHOD}

\subsection{Overview}

In our current framework, we employ the BEVDET4D \cite{bevdet4d} network as the baseline model, which is an enhanced BEV feature extraction framework. The framework consists of two main components: the backbone (including the neck) and the head. The backbone network employs ResNet \cite{ResNet} to extract both low-level geometric features and high-level semantic information. Subsequently, it utilizes the BEV generation method proposed in LSS to translate these features and semantic information into the BEV feature space, thereby obtaining BEV features. The head uses the CenterPoint \cite{centerpoint} detector head for 3D target detection. However, this method is traditionally applied to generating BEV from look-around images. By modifying the BEV generation process and introducing timestamp concepts, we establish a collaborative task framework named UVCPNet capable of time synchronization through timestamps. UVCPNet is an aerial-ground collaborative framework for enhanced BEV features that combines multi-domain information. The overall framework of the method is illustrated in Fig. 2.
\begin{figure*}
  \begin{center}
  \includegraphics[width=7 in]{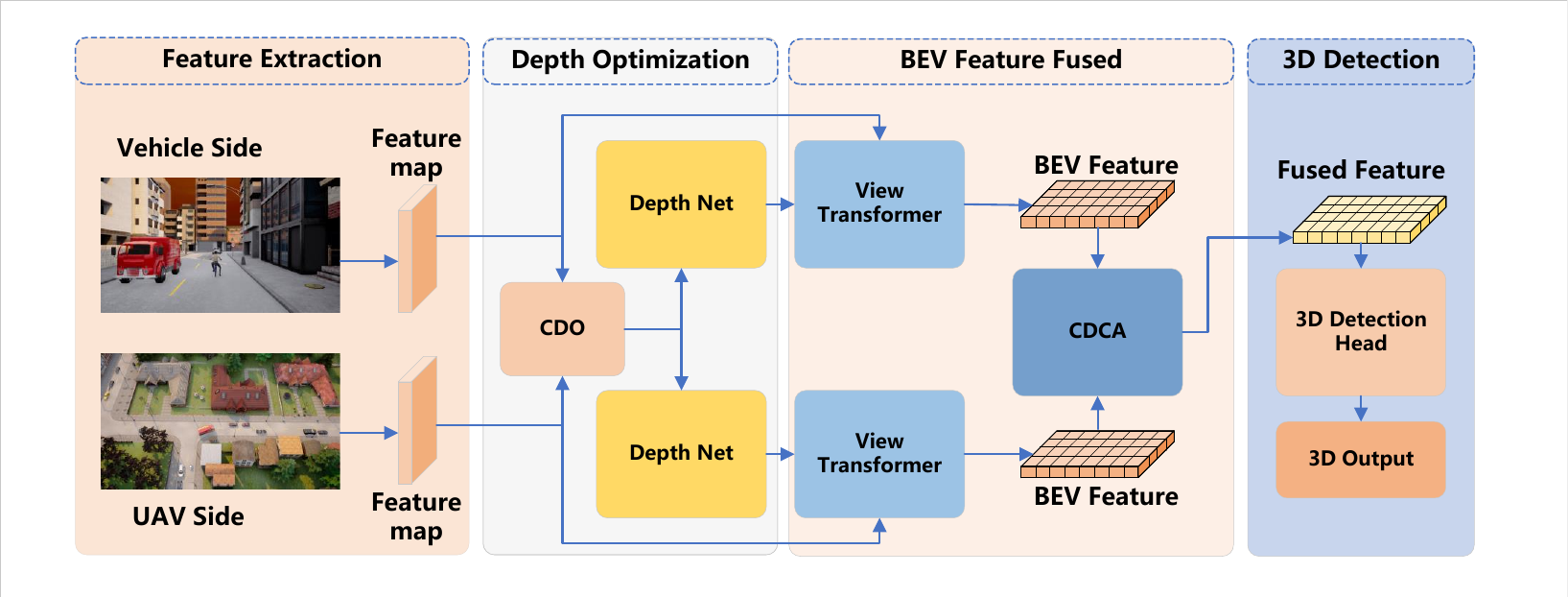}\\
  \caption{ Overview of the proposed framework. The whole collaborative inference process can be divided into four parts: 1) feature extraction: each agent
extracts the features of the input image; 2) depth optimization: get accurate depth value through collaborative optimization; 3) BEV feature fused: get aligned and enhanced Bev feature map; and 4) 3D Detection: detect 3D objects.}\label{circuit_diagram}
  \end{center}
\end{figure*}

First, we input a set of RGB images of vehicles and UAVs. Second, by combining the features extracted from these RGB images with the estimated depth information, we produce the corresponding BEV feature map for each end. It should be emphasized that the required depth information is obtained through the CDO module. The module optimizes the depth estimation network by using CRF combined with the context semantic information of two perspectives, so as to obtain accurate depth estimation results.

Third, we obtain cross-domain BEV features from multiple ends and achieve alignment and enhancement of these features by the CDCA module. This module combines feature information from the "Aerial" and "Ground" domains, facilitating effective cross-domain alignment and feature enhancement. By adjusting the weight of BEV features in each domain, we fuse the enhanced BEV feature map. Finally, we use 3D target detection method to process the fused BEV feature results, and then decode the BEV representation into 3D targets for target detection.

\subsection{CDO module}

Currently, to generate a BEV from optical images, we need to perform a 2D to 3D transformation \cite{gedepth}, represented by the following equation:

\begin{equation}
d\begin{bmatrix}
  u \\
  v \\
  1
\end{bmatrix}=K\begin{bmatrix}
 R & T \\
 0 & 1
\end{bmatrix}\begin{bmatrix}
 x_w\\
 y_w\\
 z_w\\
1
\end{bmatrix}
\end{equation}

In this context, \(d\) represents the depth value of the image, which is the distance of each pixel point from the camera's focal point. \(\mathbf{K}\) is the intrinsic matrix, \(\mathbf{R}\) is the rotation matrix of the camera, and \(\mathbf{T}\) is its translation matrix. Therefore, to obtain accurate 3D information, it is crucial to acquire precise depth estimation data.

In existing BEV-based methods aimed at addressing depth estimation inaccuracies, such as BEVdepth \cite{BEVDEPTH} and Solofusion \cite{SOLO}, additional supervision signals in the form of depth information are necessitated. This requirement imposes additional computational burdens on the agent and increases the volume of information that the agent must process. Our method leverages the contextual semantic information inherent in RGB images as implicit supervision signals to optimize the depth estimation network. This approach eliminates the need for additional supervision signals, thereby reducing the burden on agents while simultaneously enhancing depth estimation accuracy. We hope to conduct depth estimation without adding any supervisory information.

The accuracy of depth estimation plays a pivotal role in determining the precision of spatial information for targets within the BEV space \cite{philion2020lift}. Nevertheless, the majority of cooperative algorithms still rely on basic convolutional neural networks (CNNs) for depth estimation. If we consider depth estimation as a segmentation task and represent each class as a specific depth range, CRF can be utilized to enhance the quality of depth estimation. Since each domain observes the same region, the depth information for the same target should theoretically be consistent after transformation.

However, in current supervised methods, the sampling of feature maps is relatively large, causing the feature maps to lose a significant amount of adversarial information, thereby compromising the effectiveness of CRF optimization for depth. To address this issue, CRF can be employed to refine the semantic information inherent in the optical image context of both the "Aerial" and "Ground" domains. Furthermore, leveraging semantic information priors within the depth estimation network can mitigate the challenge of limited depth supervision. This approach significantly improves depth consistency at the pixel level.

We aim to use semantic information from multiple domains for comprehensive optimization. Given that each domain observes the same target, the semantic information within each domain exhibits high similarity. On the feature map of RGB image feature extraction in each domain, we let $ \{X_1^m,...,X_N^m  \}$ represent the $N$ pixels in the feature graph of the subsampled $m$ field, and $\{D_1^m,...,D_k^m\} $ denote $k$ discrete depth values. The responsibility of the depth network is to assign each pixel to various depth values, expressed as $d^m  = \{x_1^m,...,x_N^m  |xi^m\in\{D_1^m,...,D_k^m\}\}$. For depth information  $d$ within each feature map, our objective is to minimize its energy cost $E(d|s)$ \cite{Krähenbühl_Koltun_2011}:

\begin{equation}
  E(d|S)=\sum_{i}\psi_u (x_i)+\sum_{i \ne j}\psi_p(x_i,x_j) 
\end{equation}

Among these, $\psi_u (x_i)$ represents a unary potential, serving to gauge the cost associated with the initial network output:

\begin{equation}
  \sum_{i}\psi_u (x_i) = -log P(x_i|d) 
\end{equation}

$P(x_i|d)$ is the probability that the depth information of pixel $x_i$ is d. Building upon prior research \cite{CRF_PRE1,CRF_PRE2}, we define the binary potential as follows:

\begin{equation}\begin{split}
  \psi_p (x_i,x_j )=& \sum_w w\ exp(-\frac{\bar{S}_i^m-\bar{S}_j^m}{2\theta ^2 })|x_i^m- x_j^m| +\\
  & \sum_ww\ exp(-\frac{\bar{S}_i^m-\bar{S}_j^n}{2\theta ^2 })|x_i^m-x_j^n |  
\end{split}
\end{equation}

Here, $\bar{S}_i$ and $\bar{S}_j$ represent the semantic information of each image block with the same subsampling step in different domains, denoted by \( m \) and \( n \) respectively. Additionally, \( |x_i-x_j| \) signifies the compatibility between depth information, quantifying the distance between their centers in the real world. Utilizing CRF as an optimization layer acting upon the depth estimation network, we ultimately obtain the optimized depth information \( \tilde{d} \).

In essence, by employing the CDO module, we achieve more precise depth estimates for pixels without the need for additional supervision signals. Consequently, this leads to an enhanced accuracy in generating the BEV feature map for each domain.

\subsection{CDCA Module}

In our collaborative framework, the input consists of a set of RGB images containing vehicles and UAVs of varying heights. After these input optical image features are converted to BEV space, because there are significant differences between different domains in the field of view information and noticed information, it is necessary to align the information obtained across domains. To address this misalignment issue, we introduce a cross-domain cross-adaptation (CDCA) module. As shown in Fig.3, in our module, we perform collaborative cross-adaptation on the obtained BEV images to achieve a well-fused BEV feature map.

\begin{figure*}[h]
  \begin{center}
  \includegraphics[width=7 in]{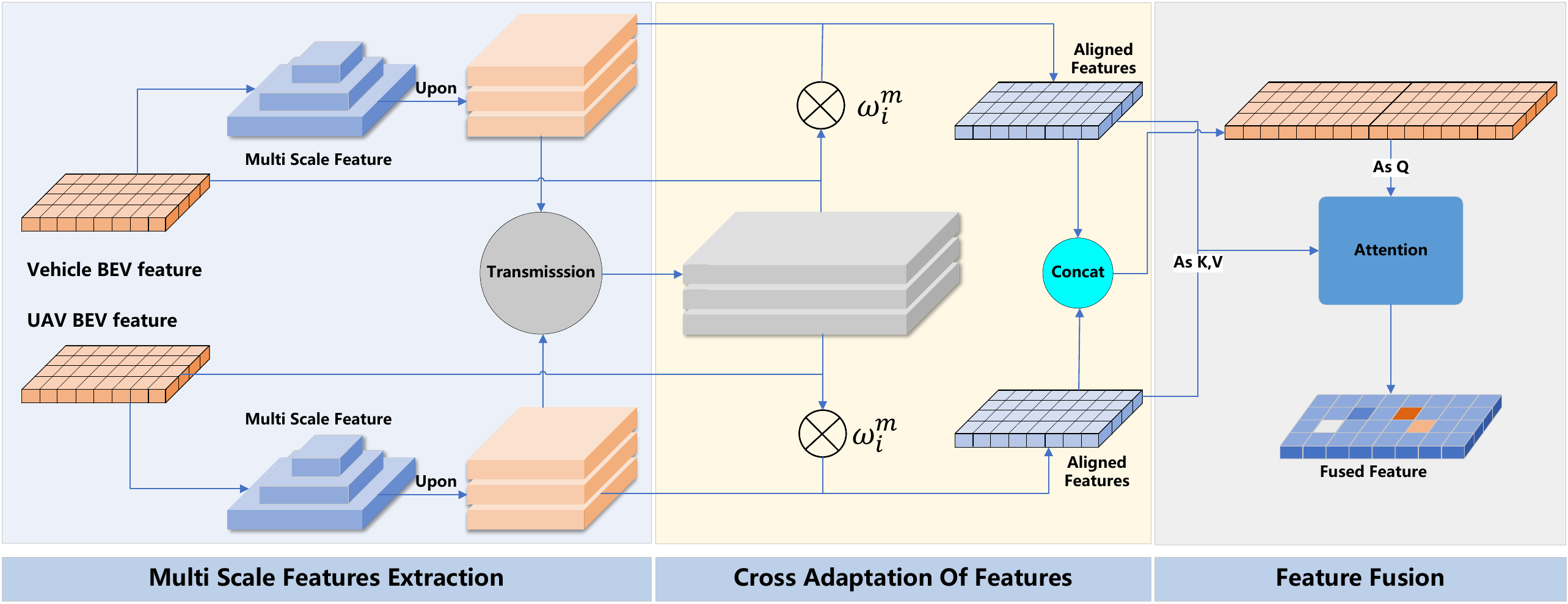}\\
  \caption{ Schematic diagram of CDCA module. It is used to align the obtained Bev feature map and fuse the Bev information at the same time.}\label{circuit_diagram}
  \end{center}
\end{figure*}

The module exploits feature correlation across domains to facilitate cross-domain adaptation, thereby enhancing feature information and minimizing positional errors. Upon acquiring the BEV feature map of each domain, it is necessary to apply consistent adaptation operations to the information from different domains in order to obtain aligned BEV feature maps. We extract multi-scale feature information from the BEV feature map of each domain, and use these multi-scale information to adapt to the feature information of other domains. After conducting correlation analysis between the information from other domains and the information from the current domain, we align and enhance the feature maps accordingly. After obtaining the enhanced feature information, we perform feature fusion to obtain the final BEV feature map.

CDCA initially requires extracting multi-scale network features. We employ a FPN \cite{fpn} structure to sample the BEV feature map $ f_i $ across four layers, resulting in the feature map $ f_i^m $ after sampling. To ensure consistency in network feature size across each scale, we utilize a Multi-Scale (MS) module to scale the feature map to the same size. After receiving the sampled feature map information $f_i^m$ transmitted from other domains, we cascade the BEV feature map information at the same level to obtain the cascaded feature map information $f^m$ at this level:

\begin{equation}
f^m=concat(f_{veh}^m,f_{uav}^m)
\end{equation}

i represents the domain of the BEV feature map, and m represents the level of the cascading feature map.

Then, we analyze the correlation of these cascaded feature maps $f^m$:

\begin{equation}
\beta _i^m=\frac{(f_i \cdot {f^m}^T)}{||f_i ||\cdot||f_i ||}\ m=(1,2,3,4)
\end{equation}

\begin{equation}
\omega _i^m=\frac{\beta _i^m}{\sum_m \beta _i^m}\ m=(1,2,3,4)
\end{equation}

\begin{equation}
f _i^{sum}=\sum_m \omega _i^m \cdot f_i^m
\end{equation}

Within this process, $\beta_i^m$ represents the autocorrelation analysis results of the features $f_i$ within the domain and the sampled feature maps within the same domain, as well as the cross-correlation analysis results between this domain and other domains. $\omega _i^m$ represents the weight information calculated through correlation analysis for each cascaded feature map, denoted as $f^m$, $f_i^sum$ represents the enhanced and aligned BEV feature map generated by fusion. By performing this operation, we can ascertain the adaptability outcomes for each scale and derive adaptability weights $\omega_i^m$ for each scale. Subsequently, we compute the enhanced feature graph information $f_i^{sum}$ within each domain.

After transforming the information from multiple domains into the BEV space, aligning and enhancing the BEV feature maps, we proceed with the fusion of BEV features. However, due to variations in the observation capabilities of each domain towards the target, ground perspectives typically capture more feature information, while aerial perspectives tend to gather relatively less target information. This leads to inconsistencies in the amount of information contained within the BEV feature maps. To address this issue, we utilize BEV space attention to fuse the BEV feature maps from both aerial and ground domains. This means that the final fused feature map will pay more attention to the BEV feature maps with higher information content, enabling them to play a more crucial role in the final fusion result, thereby enhancing the effectiveness of feature fusion.

This module enhances the model's attention to different domains by adjusting the weights of the BEV feature maps between the input domains. Firstly, the module connects the cascaded BEV feature maps of the two domains to obtain the concatenated feature map information  $K_{\text{cat}}$, $V_{\text{cat}}$  of the aerial-ground domains. Subsequently, we utilize the input feature information as $Q_{\text{veh}}$  and  $Q_{\text{uav}}$  on both sides. Finally, the weights of different domains are computed to guide the fusion process:

\begin{equation}
f _i=softmax(\frac{Q_m K_{cat}^T}{\sqrt{d_k}})V_{cat}\    (m=veh,uav)
\end{equation}

Since the output of cross-domain attention in the module is the weighted sum of $K_{\text{cat}}$, which contains all information from the two domains, the essence of the module is to bring the two domains closer to a common center point. Specifically, by leveraging the shared $K_{\text{cat}}$ in the cross-domain attention mechanism, this module encourages convergence between the two domains towards a central representation. Finally, a learnable parameter $\lambda$ is set to balance the BEV feature maps from the two domains after the attention mechanism:
\begin{equation}
f _p=\lambda \cdot f_{veh} + (1-\lambda) \cdot f_{uav}
\end{equation}

The module adjusts the ratio of the two parts of the feature maps by learning the weights in different domains. Through this operation, we can obtain a BEV fusion feature map with more accurate information.

\subsection{Loss}

The loss of the experiment is mainly composed of two parts, including smooth L1 Loss for bounding box $L_{bbox}$ and gaussian focal loss for classification $L_{cls}$ and cross-entropy loss for direction $L_{dir}$. The final loss function is as follows:

\begin{equation}
L=(\lambda_{bbox}L_{bbox}+\lambda_{cls}L_{cls}+\lambda_{dir}L_{dir})
\end{equation}

where $\lambda_{bbox}$, $\lambda_{cls}$ and $\lambda_{dir}$ represent the weights of $L_{bbox}$, $L_{cls}$ and $L_{dir}$ respectively.
\section{EXPERIMENT}

In this section, we evaluate the performance of our designed aerial-ground collaborative perception network using both a custom dataset and a publicly available dataset focused on vehicle-road collaboration. Our evaluation is structured into three parts: First, we introduce the datasets utilized in our experiments. Next, we conduct ablation studies to assess the performance of individual components of our method. Finally, we demonstrate the superiority of our proposed framework through intuitive results visualization.

\subsection{Datasets and Experimental Settings}

Currently, publicly available collaborative 3D target detection datasets are limited to low-altitude cooperative sensing tasks between vehicles and infrastructure. This gap highlights the absence of datasets for high-altitude collaborative perception involving both vehicles and UAVs. In this study, we utilize the Carla simulation software to construct the V2U-COO dataset, designed specifically for air-to-ground cooperative sensing tasks. This dataset serves as a basis for our investigation into 3D target detection within cooperative sensing tasks.

Given the limitations of existing publicly accessible collaborative 3D target detection datasets, which are confined to low-altitude vehicle and infrastructure cooperative sensing tasks, there is a notable gap in datasets for high-altitude collaborative perception involving both vehicles and UAVs. To address this, we utilized the Carla simulation software to develop the V2U-COO dataset, specifically tailored for air-to-ground cooperative sensing tasks. This dataset facilitates our investigation into 3D target detection within these cooperative sensing scenarios. Furthermore, to validate the effectiveness of our method with real-world data, we conducted experimental tests on the DAIR-V2X, an open vehicle-road collaboration dataset.

\subsubsection{DATASETS}\ 

\ \ \ \ \emph{a) V2U-COO:}
In this study, we utilized the NVIDIA RTX-3090 graphics card to construct a virtual dataset for aerial-ground cooperative sensing using the CARLA simulation software. This dataset facilitates multi-agent collaboration by using the same timestamp and identical annotation information for the same target by both vehicles and UAVs. The setup includes one vehicle and two UAVs collaborating in perception tasks. The UAVs are positioned on both the left and right sides of the vehicle to simulate accompanying flight scenarios and also to emulate different oblique viewing angles. 

\begin{figure}[h]
  \begin{center}
  \includegraphics[width=3.5 in]{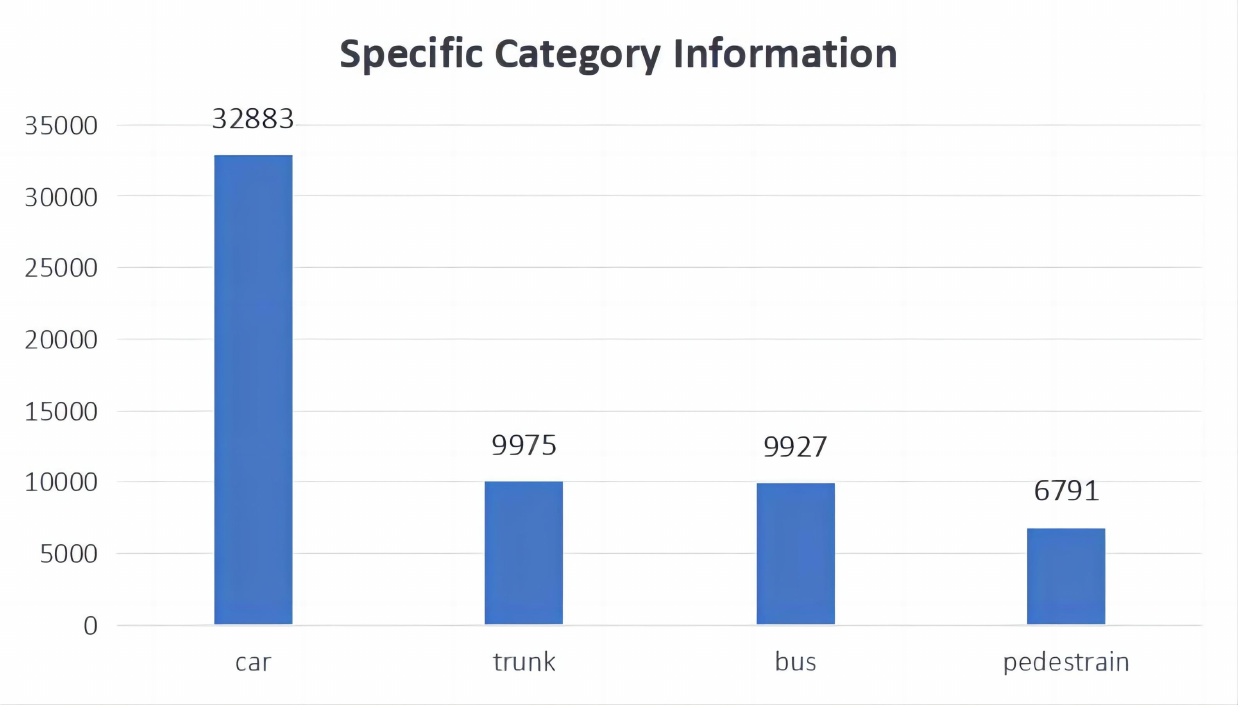}\\
  \caption{The specific category information contained in V2U-COO dataset.}\label{circuit_diagram}
  \end{center}
\end{figure}

\begin{figure*}[h]
  \begin{center}
  \includegraphics[width=7 in]{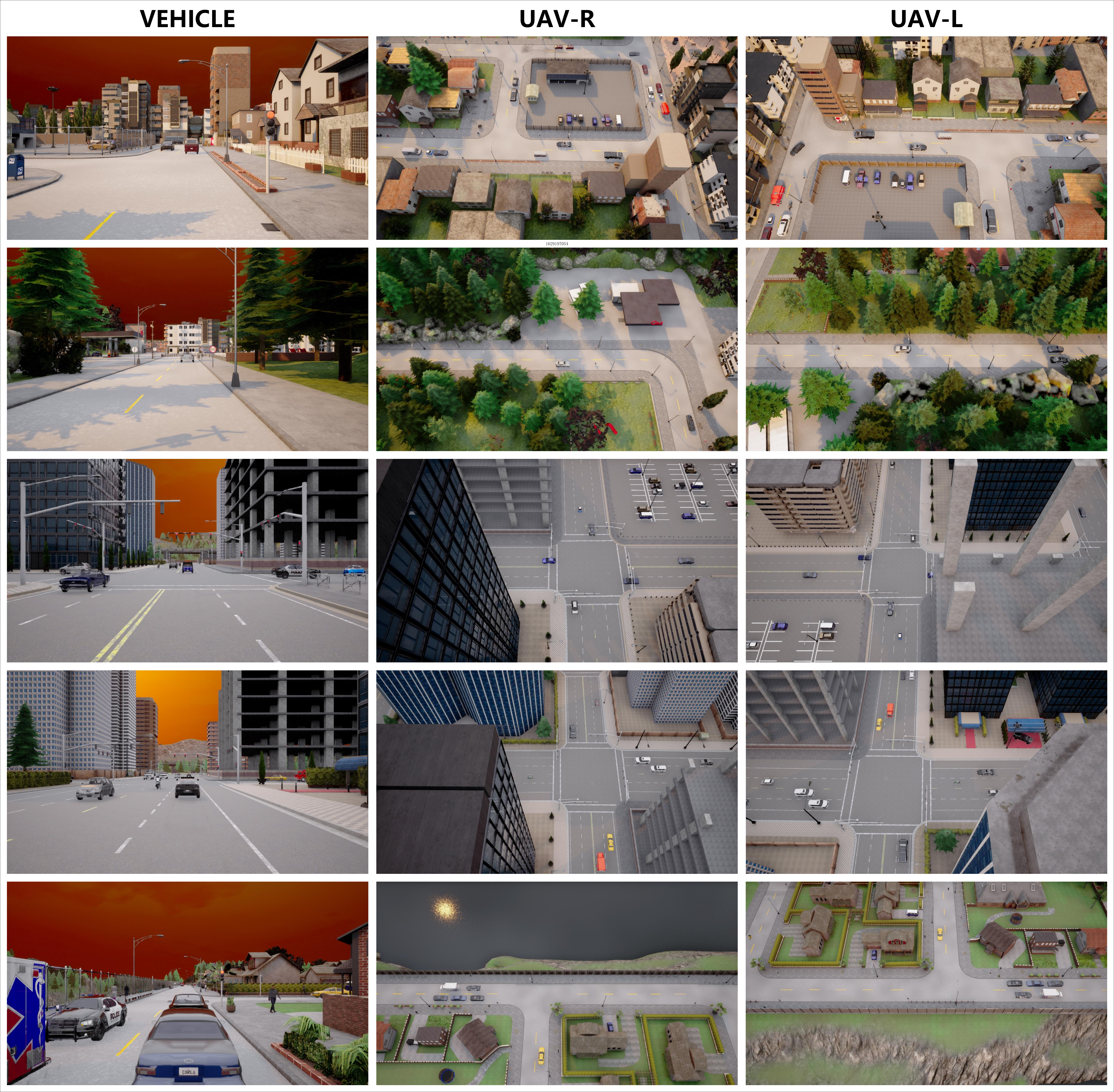}\\
  \caption{ Example diagram of multiple scenarios for v2u-coo dataset.}\label{circuit_diagram}
  \end{center}
\end{figure*}

The dataset is configured with scenarios where two UAVS are positioned at the front-left and front-right of an unmanned vehicle, with both UAVs and the vehicle observing the same area. The perception range is set at 160 meters by 110 meters. All images in the dataset are annotated with 3D objects. The dataset categorizes the objects into four classes: car, truck, bus, and pedestrian. Additionally, each agent is equipped with an optical imaging camera, enabling tasks such as target detection and tracking under cooperative optical image perception. We provide a detailed description of the specific configurations of the V2U-COO dataset in Table 1. The specific category information contained in V2U-COO dataset is shown in Fig. 4. Examples from the experimental dataset are illustrated in Fig. 5.

\begin{table}[h] \centering 
    \caption{Detailed description of the specific configurations of the V2U-COO dataset} \label{table_trois_camions}
    \begin{tabular}{cccc} \toprule 

     & {\footnotesize VEHICLE} &  {\footnotesize UAV-R} & UAV-L \\ 
     \midrule
        FOV$(^{\circ})$ & 70 & 70 & 70 \\
        Image Size & $ 1600\times 900 $  & $1600\times 900$ & $1600\times 900$ \\
        Image Count & 6400 & 6400 & 6400 \\
        Agent Height(m) & 2 & 80 & -70 \\ 
        Pitch$(^{\circ})$  & 0 & -90 & 90 \\
        Yaw$(^{\circ})$  & 0 & -60 & -60 \\
        Roll$(^{\circ})$ & 0 & 0 & 0 \\
        Sampling frequency(HZ) & 20 & 20 &20 \\
        Scene count & 7 & 7 & 7 \\
        
    \bottomrule
    \end{tabular} 
\end{table}

\ \ \ \ \emph{b) DAIR-V2X:}
DAIR-V2X is the world's first and only publicly available real dataset applied to vehicle-road cooperative perception, featuring an exceptionally large volume of data. It includes images captured by optical cameras at two different heights, from both vehicles and infrastructure. All images in this dataset are annotated with ten 3D object classes. The dataset comprises 9,311 pairs of optical images captured simultaneously from infrastructure and vehicle frames. Annotations for each pair of cooperative optical images are provided in world coordinates, which need to be transformed into vehicle coordinates for training and evaluation. These labels serve as the benchmark for our experiments. Images captured by RGB cameras have a resolution of 1920 × 1080, with a perception range of 100m × 79m.

\subsubsection{Experimental Settings}

When observations are conducted using a single agent, occlusions can occur, negatively impacting our detection and recognition capabilities. In contrast, cooperative perception allows for a more complete acquisition of sensory information, significantly reducing the effects of occlusions. Our experiments begin by confirming the effectiveness of this cooperative approach. The experimental setup integrates observational data from multiple agents to demonstrate the efficiency of the collaborative framework.

In autonomous scenarios devoid of collaborative mechanisms, observations conducted by individual agents are constrained by their inherent viewing perspectives. This limitation is particularly pronounced in observations from standalone intelligent vehicles, where conventional viewing angles frequently lead to occlusions. These occlusions present a formidable challenge in vehicle perspective observations and are identified as a critical barrier in current ground target detection methodologies. To augment the observational capabilities of vehicles, it is essential to achieve a more exhaustive acquisition of scene information. However, the restricted viewing angles often result in incomplete or non-existent data capture, especially when the intended targets are obscured by other objects. To mitigate the deficiencies caused by such occlusions, this study leverages UAV perspectives as an auxiliary observational resource. By integrating data from UAVs, the framework addresses the incomplete data issues and substantially enhances the accuracy of the perceptual information. Fig. 6 demonstrates a scenario where the vehicle’s view is obstructed; nevertheless, the integration of UAV-derived data facilitates a comprehensive scene understanding through collaborative perception.

\begin{figure}[h]
  \begin{center}
  \includegraphics[width=3.5in]{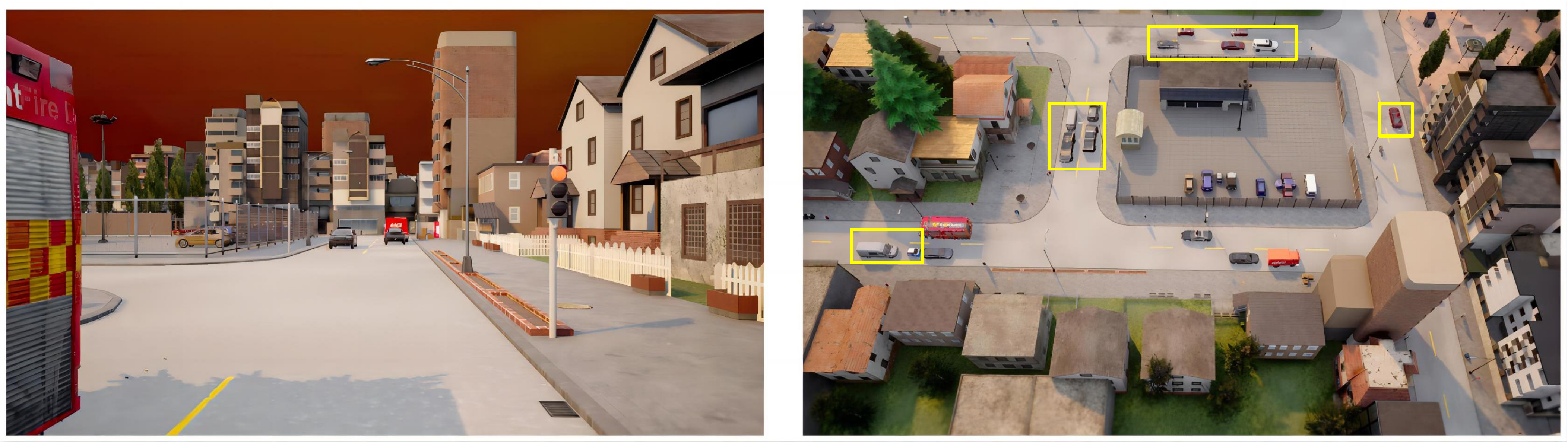}\\
  \caption{UAV observes the area where the vehicle's viewing angle is blocked.}\label{circuit_diagram}
  \end{center}
\end{figure}

When observations are conducted solely using UAVs, the significant observation distance often results in a lack of detail in the visual data captured, particularly for smaller objects. Consequently, the detection accuracy for smaller targets presents a substantial challenge in UAV-based monitoring systems. To enhance the detail and accuracy of the data collected during UAV operations, it is advantageous to integrate these observations with data obtained from ground-based vehicle sensors. Such collaborative sensing can significantly improve the fidelity of the detail captured, leading to more accurate scene perception. An illustrative example is provided in the Fig. 7, which depicts a scenario where a vehicle captures detailed information about a small target initially observed by a UAV.

\begin{figure}[h]
  \begin{center}
  \includegraphics[width=3.5 in]{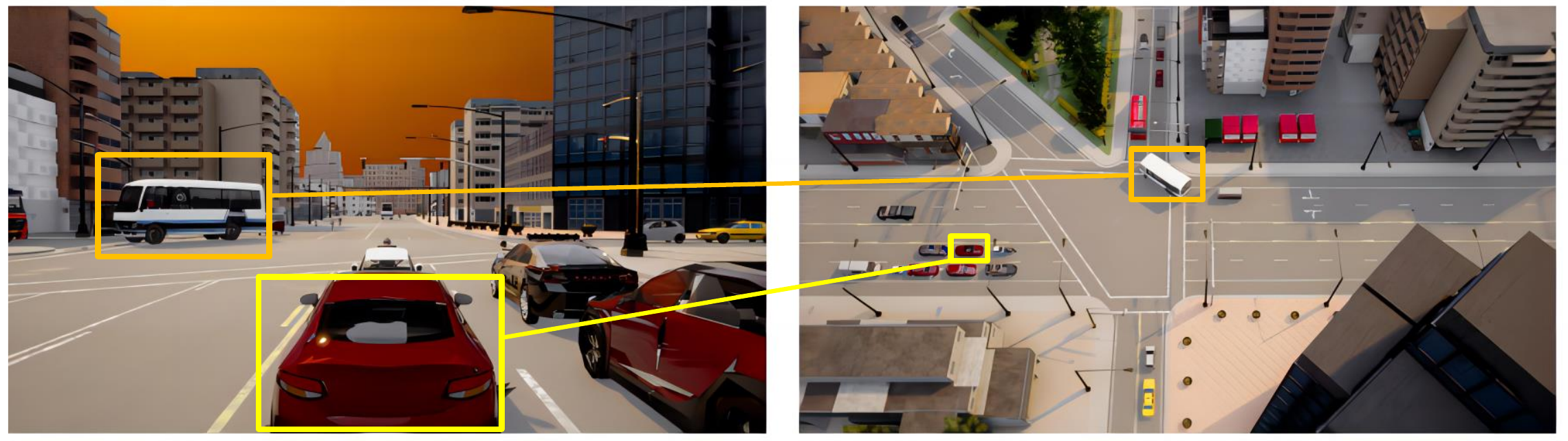}\\
  \caption{Detailed information of vehicle observation complements UAV observation.}\label{circuit_diagram}
  \end{center}
\end{figure}

Therefore, by synergizing the observational data from both vehicles and UAVs, a more comprehensive understanding of the target can be achieved, enhancing the overall accuracy of the information gathered.

\subsection{Evaluation Metrics}

We evaluate the performance of the network using the evaluation metrics provided by Nuscenes \cite{nuscenes2019}. These metrics include the mean average precision (mAP), mean translation error (mATE), mean scale error (mASE), mean orientation error (mAOE), mean velocity error (mAVE), mean attribute error (mAAE), and the NuScenes detection score (NDS), which considers both comprehensive accuracy and true positives (TP).

The average precision of object detection is a crucial metric for assessing detection accuracy. However, in NuScenes, the calculation of mAP integrates threshold matching of average precision (AP) without employing Intersection over Union (IoU). Instead, it uses the 2D center distance (d) on the ground plane to compute AP, thereby decoupling the impact of object size and orientation on AP calculation. Here, d is set to $D=\{0.5,1,2,4\}$ meters. The formula for mAP calculation is as follows:

\begin{equation}
mAP=\frac{1}{\left|\mathbb{C}\right|\left|\mathbb{D}\right|}\sum_{c\in\mathbb{C}}\sum_{d\in\mathbb{D}}{AP}_{c,d}
\end{equation}

For each TP index, we calculate as follows:
\begin{equation}
mTP = \frac{1}{\mathbb{C}}\sum_{c\in\mathbb{C}}{TP_c}
\end{equation}

NDS is computed using the TP metric, where half is based on mAP, and the other half is based on the quality of detection performance in terms of position, size, orientation, attributes, and velocity (ATE, ASE, AOE, AVE, AAE). As mAVE, mAOE, and mATE can exceed 1, we constrain each metric between 0 and 1 in the following formula:
\begin{equation}
NDS =\frac{1}{10}[5mAP+\sum_{TP\in \mathbb{TP}}{(1-mTP)}]
\end{equation}

\subsection{Implementation Details}

The framework proposed in this study is experimented with using PyTorch on an RTX-4090 GPU to evaluate its efficacy. To assess the effectiveness of the proposed collaborative framework, we ensure uniformity by employing the same training detection model architecture across all agent nodes. We opt for ResNet-50 as the pre-trained model and subsequently train it on respective datasets.

During the training phase, parameters remain static at each agent, ensuring that updates occur solely within the collaborative module. This guarantees that all improvements stem from the collaborative framework, facilitating a comprehensive evaluation of the network's capabilities. In the training detection phase, input images are standardized to 704x256 dimensions, with training extending over 40 epochs. We employ a standard gradient optimizer with a weight decay of 0.0005. Additionally, the learning rate is initialized at 0.001.

\subsection{Ablation Study}

In this section, we conduct experiments based on the proposed V2U-COO dataset. Firstly, to assess the effectiveness of the collaborative network, we compare the results obtained from using only vehicle observations or only UAV observations with those obtained from collaborative network observations under the same network architecture. The specific mAP improvement curve is illustrated in the figure, demonstrating that the collaborative platform outperforms both the vehicle-only and UAV-only approaches. Additionally, to evaluate the effectiveness of each module within UVCP, we conduct extensive experiments on the dataset. To ensure fairness, experiments are conducted in the same environment. The overall experimental results are summarized in the Table II. It can be observed that compared to the baseline model, UVCP achieves an average precision improvement of approximately 6.1\%.

\begin{table*}[h] \centering 
    \caption{The overall experimental results} 
    \label{table_trois_camions}
    \resizebox{1.5\columnwidth}{!}{
    \begin{tabular}{c|ccccccc|c} \toprule 

     & mAP & mATE & mASE & mAOE & mAVE & mAAE & NDS & mAP$\Delta$ \\ 
     \midrule
        Baseline- & 0.546 & 0.380 & 0.167 & 0.564 & 1.157 & 0.387 & 0.523 &  \\
        Baseline+ & \textbf{0.607} & 0.355 & 0.145 & 0.544 & 1.082 & 0.315 & 0.568 & \textbf{6.1\%$\uparrow$} \\
    \bottomrule
    \end{tabular} }
\end{table*}

\subsubsection{Effectiveness analysis of collaborative perception framework}

Table III presents the experimental results conducted on the V2U-COO dataset, comparing the outcomes of utilizing a single agent versus collaborative observation between agents and reverse observation.,The specific accuracy improvement curves are depicted in the Figure 3. When employing single-agent observation, the average precision achieved solely through vehicle observation is merely 12.1\%, whereas the average precision attained solely through UAV observation stands at 44.7\%. However, upon integrating the collaborative perception framework, the average precision of collaborative detection significantly improves to 60.7\%. This improvement is substantial compared to single-agent observations, with a remarkable 48.6\% enhancement over vehicle observation and a notable 16.0\% improvement over UAV observation.

\begin{figure}[h]
  \begin{center}
  \includegraphics[width=3.5 in]{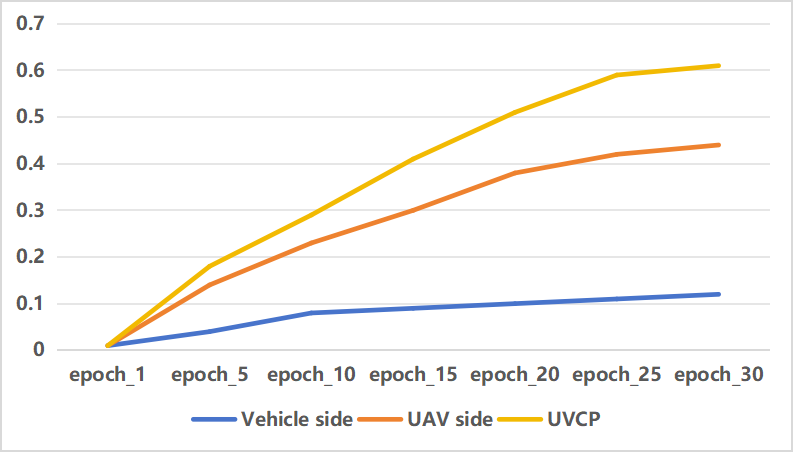}\\
  \caption{Accuracy curves of single agent and collaborative framework.}\label{circuit_diagram}
  \end{center}
\end{figure}

\begin{table*}[h] \centering 
    \caption{Accuracy results of single agent and collaborative framework}
    \label{table_trois_camions}
    \resizebox{1.5\columnwidth}{!}{
    \begin{tabular}{c|ccccccc|c} \toprule 

     & mAP & mATE & mASE & mAOE & mAVE & mAAE & NDS & mAP$\Delta$ \\ 
     \midrule
        Vehicle side & 0.121 & 0.855 & 0.366 & 0.677 & 1.358 & 0.467 & 0.224 & \textbf{48.6\%$\downarrow$} \\
        UAV side     & 0.447 & 0.562 & 0.187 & 1.375 & 2.108 & 0.511 & 0.397 & \textbf{16.0\%$\downarrow$} \\
    \midrule
        UVCP         & \textbf{0.607} & 0.355 & 0.145 & 0.544 & 1.082 & 0.315 & 0.568 &  \\
    \bottomrule
    \end{tabular} }
\end{table*}

In the aforementioned experiments, we observed that due to occlusion and limited field of view, the results of single-agent vehicle observation were unsatisfactory. Similarly, the effectiveness of UAV observation was compromised due to the scarcity of target information. However, when we integrated these two platforms into a collaborative perception system, a significant improvement in accuracy was observed. This validates the effectiveness of collaborative perception and demonstrates that UAVs can provide a broader field of view for vehicles, while vehicles can offer more precise target information for UAVs.
\subsubsection{Effectiveness analysis of each module in collaborative perception framework}

As described in Section III, our proposed framework comprises two main modules: the cross-domain adaptation module and the depth estimation optimization module. We conducted a series of ablation experiments on the V2U-COO dataset to analyze the importance of each module.

\begin{figure}[h]
  \begin{center}
  \includegraphics[width=3.5in]{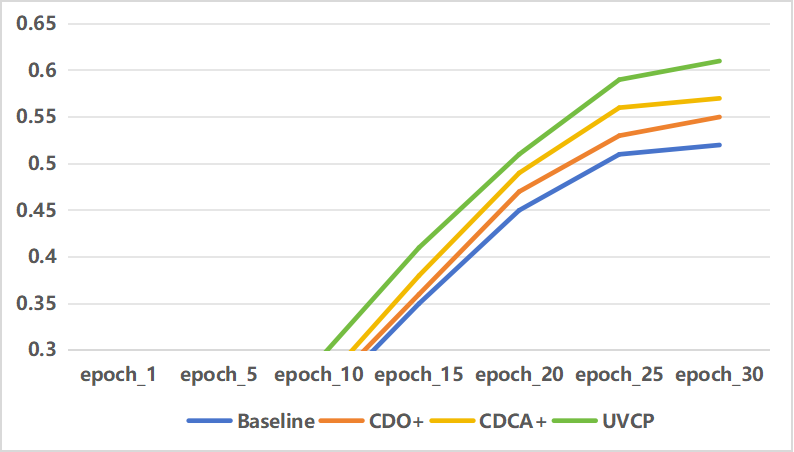}\\
  \caption{Accuracy curves of different components in the UVCP.}\label{circuit_diagram}
  \end{center}
\end{figure}

\begin{figure*}[!t]
  \begin{center}
  \includegraphics[width=1.0\textwidth]{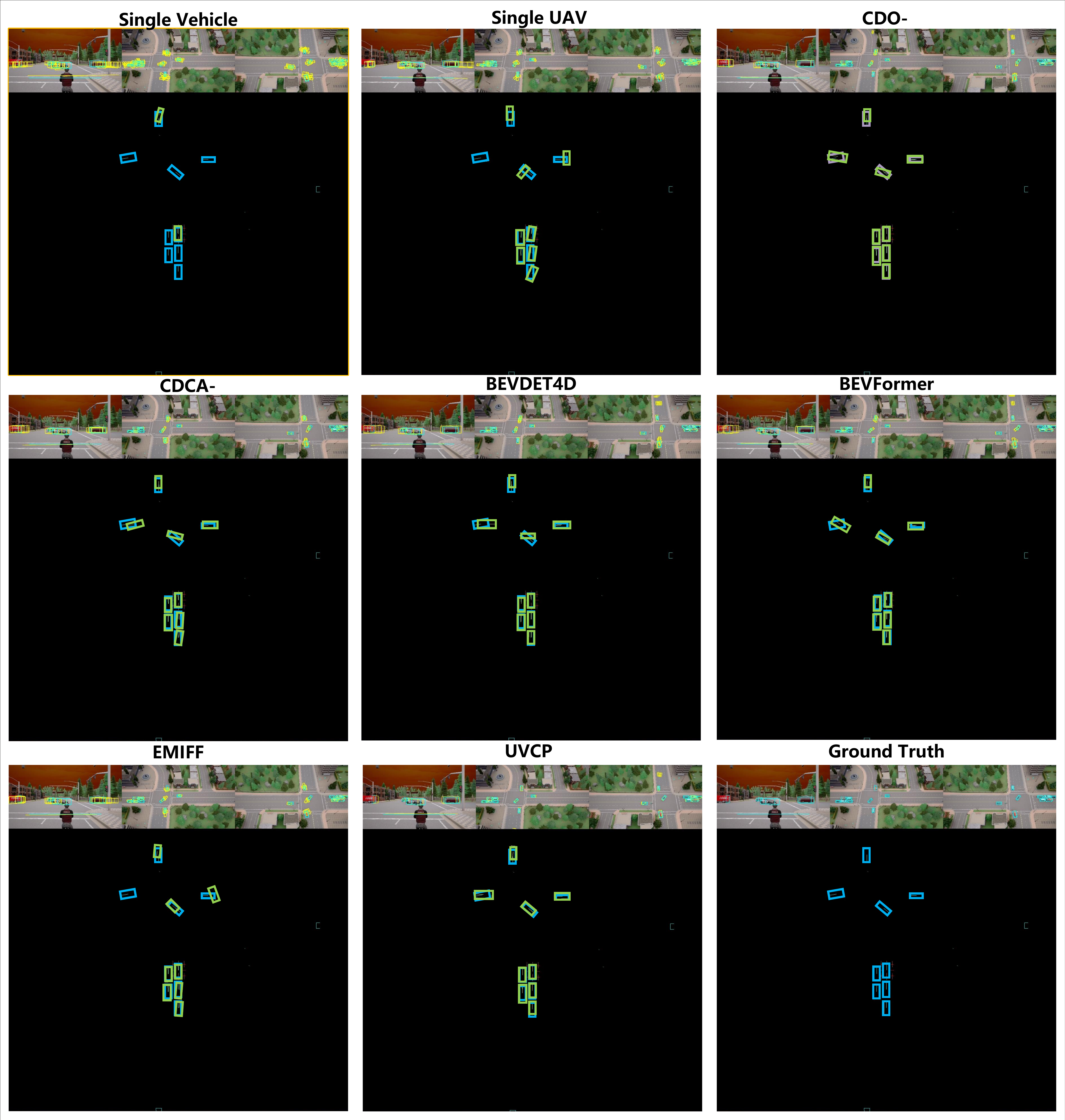}\\
  \caption{Visualization results. The upper part is the result in the optical image, and the lower part is the result in the BEV diagram. The blue box is the true value, and the yellow box is the detection result.}\label{circuit_diagram}
  \end{center}
\end{figure*}

\ \ \ \ \emph{a) Effectiveness analysis of CDO module:}
The framework relies on accurate depth values during the generation of BEV feature maps. Therefore, achieving precise depth estimation is a crucial aspect of the collaborative perception framework. To address this challenge, we propose the depth estimation optimization module, which leverages contextual semantic information to obtain refined and accurate depth values.

\begin{table*}[h] \centering 
    \caption{Ablation results of CDO module} 
    \label{table_trois_camions}
    \resizebox{1.5\columnwidth}{!}{
    \begin{tabular}{c|ccccccc|c} \toprule 

     & mAP & mATE & mASE & mAOE & mAVE & mAAE & NDS & mAP$\Delta$ \\ 
     \midrule
        CDO- & 0.576 & 0.377 & 0.151 & 0.423 & 0.954 & 0.352 & 0.561 &  \\
        CDO+ & \textbf{0.607} & 0.355 & 0.145 & 0.544 & 1.082 & 0.315 & 0.568 & \textbf{3.1\%$\uparrow$} \\
    \bottomrule
    \end{tabular} }
\end{table*}

The ablation experiments for this module are presented in Table IV, where we compare the changes in various metrics of the model before and after depth optimization, denoted as CDO+ and CDO-, respectively. The results demonstrate that after integrating this module, the model's accuracy improves from 57.6\% to approximately 60.7\%, resulting in an enhancement of about 3.1\% in accuracy.

Since the depth estimation network performs contextual semantic optimization on images downsampled to a size of 704 × 256, we further validate the effectiveness of the module by reducing the downsampling factor of images to obtain sizes of 1056 × 384 and 1408 × 512 for depth optimization. The experiment results, as shown in Table V, confirm the effectiveness of depth optimization.

\begin{table*}[bhtp] \centering 
    \caption{Effectiveness analysis of CDO module} 
    \label{table_trois_camions}
    \resizebox{1.5\columnwidth}{!}{
    \begin{tabular}{c|ccccccc|c} \toprule 

    Input size & mAP & mATE & mASE & mAOE & mAVE & mAAE & NDS & mAP$\Delta$ \\ 
     \midrule
        704 × 256  & 0.607 & 0.355 & 0.145 & 0.544 & 1.082 & 0.315 & 0.568 &  \\
        1056 × 384 & 0.618 & 0.345 & 0.167 & 0.678 & 1.308 & 0.289 & 0.564 & \textbf{1.1\%$\uparrow$} \\
        1408 × 512 & \textbf{0.631} & 0.334 & 0.139 & 0.593 & 1.493 & 0.343 & 0.575 & \textbf{2.4\%$\uparrow$} \\
    \bottomrule
    \end{tabular} }
\end{table*}

\ \ \ \ \emph{b) Effectiveness analysis of CDCA module:}
Our primary task in collaborative perception is to integrate information observed by multiple agents into a cohesive representation. However, the challenge lies in aligning the disparate information captured by these agents, leading to inaccurate target localization and redundant detections. To tackle this issue, we introduce a cross-domain cross-adaptation module. This module extracts multi-scale information from each agent and performs cross-adaptation with information obtained by other agents to derive the optimal fused information.

\begin{table*}[h] \centering 
    \caption{Ablation results of CDCA module} 
    \label{table_trois_camions}
    \resizebox{1.5\columnwidth}{!}{
    \begin{tabular}{c|ccccccc|c} \toprule 

     & mAP & mATE & mASE & mAOE & mAVE & mAAE & NDS & mAP$\Delta$ \\ 
     \midrule
        CDCA- & 0.559 & 0.329 & 0.169 & 0.409 & 1.026 & 0.372 & 0.551 &  \\
        CDCA+ & \textbf{0.607} & 0.355 & 0.145 & 0.544 & 1.082 & 0.315 & 0.568 & \textbf{4.8\%$\uparrow$} \\
    \bottomrule
    \end{tabular} }
\end{table*}

The results of ablation experiments for this module are summarized in Table VI, where we compare metrics with and without its utilization, denoted as CDCA+ and CDCA-, respectively. It is evident that the integration of this module enhances the average precision from 55.9\% to 60.7\%, marking an improvement of approximately 4.8\%. This underscores the effectiveness of the cross-domain cross-adaptation module in aligning information from diverse domains, thereby mitigating detection errors.

Furthermore, by visually comparing the feature map results of vehicles and UAVs before and after cross-domain fusion, and juxtaposing them with the fused detection BEV results, we can intuitively observe the refinement achieved through fusion.

\subsection{ Comparison With Other Methods}

To rigorously assess the efficacy of our proposed approach, we conducted a series of comparative experiments against both established BEV-based methodologies and select collaborative techniques. Our evaluations included comparisons with BEV-based systems such as BEVDet4D, BEVerse, and BEVFormer, alongside the collaborative approach EMIFF. Detailed outcomes of these comparisons are delineated in Table VII. The results demonstrate a significant enhancement in average precision with our method, exhibiting improvements of approximately 9\% over BEVDet4D, 11\% over BEVFormer \cite{bevformer}, and 8\% over BEVerse \cite{beverse}. Additionally, when compared to the collaborative method EMIFF \cite{emiff}, our approach showed an improvement of about 6\%. These findings underscore the superior performance of our proposed method in terms of average precision, highlighting its robustness and effectiveness in the tested scenarios.

\begin{table*}[bhtp] \centering 
    \caption{Comparative experiment with other methods} 
    \label{table_trois_camions}
    \resizebox{1.5\columnwidth}{!}{
    \begin{tabular}{c|ccccccc} \toprule 

    Method & mAP & mATE & mASE & mAOE & mAVE & mAAE & NDS  \\ 
     \midrule
        BEVDet4D  & 0.515 & 0.421 & 0.157 & 0.430 & \textbf{0.913} & 0.376 & 0.528   \\
        BEVerse   & 0.521 & 0.409 & 0.156 & \textbf{0.426} & 0.922 & 0.378 & 0.532  \\
        BEVFormer & 0.487 & 0.391 & 0.171 & 0.513 & 1.057 & 0.380 & 0.498  \\
        EMIFF     & 0.546 & 0.320 & 0.154 & 0.631 & 1.088 & 0.449 & 0.518  \\
        Where2com & 0.555 & \textbf{0.307} & 0.146 & 0.467 & 1.112 & 0.407 & 0.545  \\
        UVCP(ours)& \textbf{0.607} & 0.355 & \textbf{0.145} & 0.544 & 1.082 & \textbf{0.315} & \textbf{0.568}  \\
    \bottomrule
    \end{tabular} }
\end{table*}

\subsection{Results on the Difficult Dataset DAIR-V2X}

To assess the efficacy of our cooperative framework and evaluate its performance within real-world contexts, we conducted validation experiments on the DAIR-V2X dataset. This dataset, comprising a rich array of images from real-world scenarios, presents significant challenges due to various adverse conditions that inherently increase the complexity of detection tasks. The experimental outcomes, as presented in Table VIII, indicate that the application of our proposed method led to an approximate increase of 2.7\% in average precision. These results not only confirm the effectiveness of our framework but also demonstrate its robustness and adaptability in realistic settings.

\begin{table*}[h] \centering 
    \caption{Experiment on dair-v2x} 
    \label{table_trois_camions}
    \resizebox{1.5\columnwidth}{!}{
    \begin{tabular}{c|ccccccc|c} \toprule 

     & mAP & mATE & mASE & mAOE & mAVE & mAAE & NDS & $\Delta$ \\ 
     \midrule
        Baseline- & 0.336 & 0.681 & 0.185 & 0.645 & 1.095 & 0.334 & 0.368 &  \\
        Baseline+ & \textbf{0.363} & 0.657 & 0.376 & 0.676 & 1.034 & 0.344 & 0.376 & \textbf{2.7\%$\uparrow$} \\
    \bottomrule
    \end{tabular} }
\end{table*}

\subsection{Visual Analysis}
In this section, we present a comparative analysis of the visualization results obtained from individual observations by vehicles and UAVs, as well as those enhanced through the incorporation of our collaborative framework. The analysis reveals that observations from single agents often suffer from challenges such as missed or duplicate detections. However, by integrating the collaborative framework, the observational outcomes for each agent are significantly improved. This enhancement is evident across multiple scenarios, confirming the effectiveness of the collaborative approach in providing more accurate and reliable detection results.

\section{CONCLUSION}

In this paper, we propose an aerial-ground collaborative perception framework named UVCP. UVCP primarily focuses on the collaborative fusion of information observed by agents at different altitudes. Compared to scenarios where observations are made solely by vehicles or solely by drones, UVCP unifies the observational data from both drones and vehicles, addressing the limitations inherent to each single-agent perspective. This integration compensates for the deficiencies of single-view observations, thereby enhancing the accuracy of 3D object detection.

Additionally, since we convert the original images to BEV images, we designed the CDO module to ensure better depth information values and thereby obtain accurate BEV feature maps. Moreover, considering the challenge of aligning target information obtained by different agents during the fusion process, we devised the CDCA module to enable effective cross-domain information fusion through mutual adaptation of information from different perceptual perspectives.

To facilitate the research on aerial-ground collaboration, we utilized CARLA to create a dataset aimed at aerial-ground collaboration, named V2U-COO. We conducted extensive experiments and analyses on both our dataset and the publicly available DAIR-V2X dataset to verify the accuracy of our method in 3D object detection. Experimental results indicate that the detection accuracy of the collaborative strategy significantly surpasses that of single-agent scenarios. Additionally, the CDCA and CDO modules have also significantly improved the detection accuracy. However, the proposed method still has limitations in scenarios where multi-agent observations are temporally asynchronous. In our future research, we plan to address the temporal asynchrony issues by modeling historical information to reduce information errors. Additionally, we aim to leverage multi-modal data from optical and point cloud sensors in collaborative perception to further enhance the processing capabilities in distributed scenarios, which can be applied to real-world applications demanding quick response and high precision.







\bibliographystyle{IEEEtran}
\bibliography{IEEEabrv,Bibliography}

\vfill


\end{document}